\documentclass[lettersize,journal]{IEEEtran}
\usepackage{amsmath,amsfonts}
\usepackage{algorithmic}
\usepackage{algorithm}
\usepackage{array}
\usepackage{xcolor}
\usepackage{textcomp}
\usepackage{stfloats}
\usepackage{url}
\usepackage{verbatim}
\usepackage{graphicx,float}
\usepackage{authblk}
\usepackage{cite}
\usepackage{caption}
\usepackage{subcaption}
\usepackage{multicol}
\usepackage{multirow}
\usepackage[hidelinks]{hyperref}
\usepackage{threeparttable}
\begin{document}

\title{UAV Position Estimation using a LiDAR-based 3D Object Detection Method }

\author[1]{Uthman Olawoye}

\author[1]{Jason N. Gross}
\affil[1]{Department of Mechanical and Aerospace Engineering \\
West Virginia University \\
Morgantown, USA}




\maketitle

\begin{abstract}
This paper explores the use of applying a deep learning approach for 3D object detection to compute the relative position of an Unmanned Aerial Vehicle (UAV) from an Unmanned Ground Vehicle (UGV) equipped with a LiDAR sensor in a GPS-denied environment. This was achieved by evaluating the LiDAR sensor's data through a 3D detection algorithm (PointPillars). The PointPillars algorithm incorporates a column voxel point-cloud representation and a 2D Convolutional Neural Network (CNN) to generate distinctive point-cloud features representing the object to be identified, in this case, the UAV. The current localization method utilizes point-cloud segmentation, Euclidean clustering, and predefined heuristics to obtain the relative position of the UAV. Results from the two methods were then compared to a reference truth solution.  
\end{abstract}


\section{Introduction}
Light Detection and Ranging (LiDAR)-based object detection methods have drawn sufficient attention from academia and industry due to LiDAR being a sensing capability that creates a 3D representation of a surveyed environment using light in the form of a pulsed laser. Several industries, including automotive, robotics, infrastructure, and the military, use LiDAR for mapping, navigation, and other uses~\cite{molebny2016laser,sentech_2022,velodyne_lidar_2022}. These approaches offer strong performance in various lighting and weather conditions because LiDAR generates its own light. A LiDAR sensor emits pulsed light waves from a laser source into the environment, and these pulses are bounced off surrounding objects. The distance traveled by each pulse can be calculated using the time it takes to return to the sensor. A real-time 3D map of the environment can be generated by repeating this process millions of times per second. 

Like a camera image, the LiDAR point cloud can also be parsed through detection methods to identify and classify objects. Object detection in a 2D image plane is a well-studied topic, and recent advances in deep learning have demonstrated remarkable success in real-time applications \cite{liu2016ssd, ren2015faster,bochkovskiy2020yolov4}. Deep learning is a powerful technology that can automatically learn features from data, widely used to improve image-based 2D detection performance. Research into these methods has increased significantly since the KITTI dataset \cite {geiger2013vision} was published in 2012, and a significant number of methods have been proposed to learn 3D geometric features and topological structure from LiDAR point clouds and to predict object 3D location and classification in an end-to-end network~\cite{wu2020deep}. A LiDAR-based 3D object detector is attractive in that it uses the LiDAR sensor data solely without relying on data from any other sensor.       

\section{Background Problem}
LiDAR is mainly used in mobile robotics for localization and navigation purposes as, for example, ~\cite{asvadi20163d,gross2019field}.

The presented paper utilizes the data from the robot system described and implemented within ~\cite{gross2019field, de2021planning}, which includes an Unmanned Ground Vehicle (UGV), that is a drive chassis Husky by Clearpath Robotics ~\cite{clearpath_husky_a300}. The primary sensors carried by
the UGV for a cooperative mission in tandem with a quadrotor drone is a Velodyne VLS-128 channels 3D LiDAR, used for SLAM purposes, an UltraWideBand (UWB) (Decawave DWM-1001) ranging radio, and a FLIR camera with a fisheye lens. The UGV, equipped with heavier sensors and higher capacity batteries than the drone, has the task of fusing the sensor data, including the one sent by the UAV, and providing the drone poses along with the planning algorithm as described in~\cite{de2021planning}.

The UAV is a custom frame-build equipped with  PixHawk4 (PX4) flight controller, onboard computer, UltraWideBand (UWB) (DWM-1001) ranging radio that is paired with the UGV's UWB, and a LidarLite laser altimeter for altitude measurements. Figure \ref{fig:2F} shows the setup for the UAV and UGV.


\begin{figure}[H]
    \centering
    \includegraphics[scale=0.70,width=1.0\linewidth]{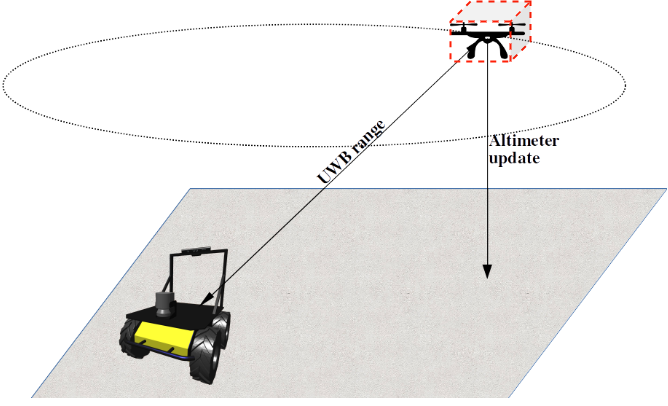}\
    \caption{\textrm{Position estimation using the clustering method}}
    \label{fig:2F}
\end{figure}

In this configuration, the LiDAR data is used to compute the relative position estimate between the UAV and UGV. To obtain the position of the UAV in the point cloud, in our prior work described ~\cite{asvadi20163d,gross2019field}, the pointcloud of the UGV is segmented into a hollow sphere whose inner and outer radii bound the UWB ranging by two meters. Next, the segmented pointcloud is passed to a fast simple k-means clustering technique, which identifies potential clusters. Available clusters in each pointcloud are then evaluated against simple heuristics to determine if a particular cluster represents the UAV. The heuristics used to distinguish a cluster as the UAV include:
\begin{itemize}
    \item Height of the pointcloud is within a manually set threshold of the UAV's onboard laser altimeter.
    \item Minimum volume constraint with respect to the bounding box defined by the extremities of the cluster
    \item Shape constraints on the bounding box that is defined by the cluster. For example, defining the allowable difference between the length, width, and height of the bounding box.
\end{itemize}

\begin{figure}[h!]
     \centering
     \includegraphics[scale = 0.25]{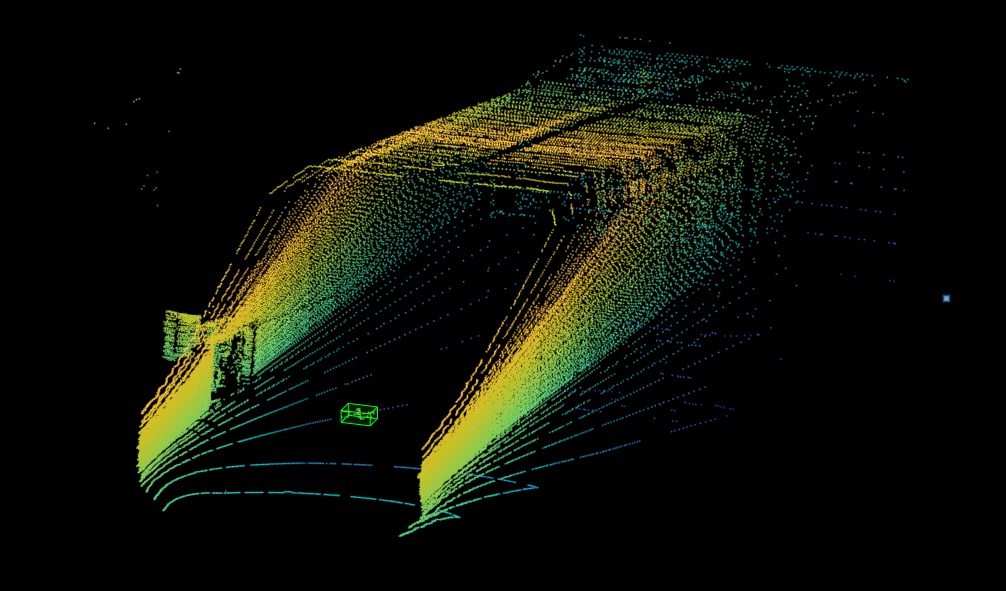}
     \caption{Pointcloud representation of the tunnel with the UAV bounded by the green box}
     \label{fig:annoPC}
\end{figure}

An example of the identified drone is shown in Fig. \ref{fig:annoPC}.

The centroid of the bounding box fitted around a cluster that passes these set of heuristics is returned as the UAV position estimate. The setback of this method is that the probability of multiple objects in the segmented pointcloud satisfying the constraints set by the heuristics is high, and the method provides no way to verify if the selected cluster is the UAV. This would, in turn, provide wrong estimates for the position of the UAV.

\begin{figure}[H]
    \centering
    \includegraphics[width=\columnwidth]{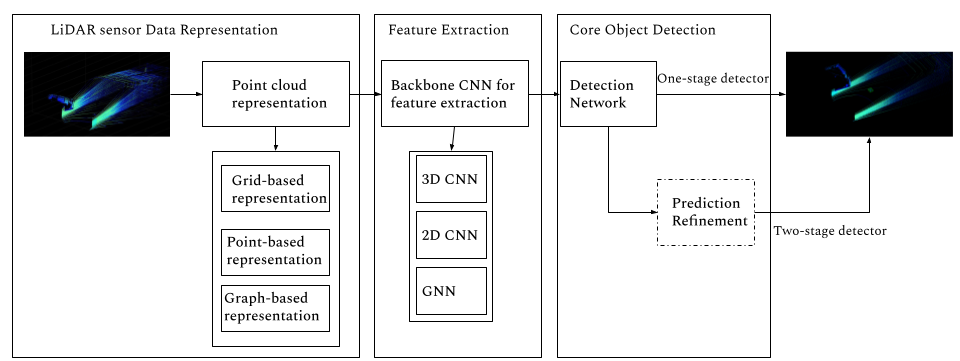}
    \captionsetup{justification=centering}
    \caption{\textrm{LiDAR-based 3D object detection pipeline}}
    \label{fig:2G}
\end{figure}

Multiple approaches were investigated to address the limitations of the simple clustering method discussed above. One of these approaches included comparing the cluster from the current epoch to the previously observed solution. Figure \ref{fig:2G} details the algorithm used in this method. The algorithm assumes the maximum velocity of the UAV should not be greater than a pre-set threshold (e.g.,  0.5 m/s). The main challenge encountered in this approach is that if the algorithm cannot find a cluster representing the UAV, the gap in time would often make the thresholding useless. Thus, with a velocity threshold, data gaps could lead to an algorithm ignoring every subsequent solution for the rest of the flight sequence. 

Pointcloud registration techniques were also considered to improve the localization solution. Such as Iterative Closest Point (ICP) [16], Normal Distributions Transform (NDT) [17], and TEASER++ [18]. These pointcloud registration techniques are combined with the clustering method to provide a verification layer for the published cluster. In this case, we assume a 3D model of the UAV is placed precisely where the LiDAR sensor is in the system. The published cluster would then be matched to the 3D model, and the resulting transformation, which provides a position estimate, is compared to the current position estimate. A critical challenge of this method is that the LiDAR sensor captures roughly
50- 70 points representing the UAV in real time. Matching a sparse, partially occluded scan to a dense 3D model becomes
complicated and inefficient.


\section{LiDAR-based 3D Object Detection}

LiDAR-based 3D Object detection methods utilize only LiDAR sensor data for inference and detection. DCNNs for feature encoding, extraction, and prediction have been simplified into a single operational pipeline by Georgios et al. \cite{zamanakos2021comprehensive}. The pipeline consists of three modules: point cloud representation, feature extraction, and the core object detection module.

\subsection{Pointcloud Representation Module}
This module is responsible for transforming and encoding the unorganized data from the LiDAR sensor into a compact structure. These representation methods can be classified into three main parts. Grid-based methods that convert irregular point clouds into regular representations. The 3D space is divided into cells of a fixed size, and feature vectors are extracted. 2D or 3D convolutional networks can process the resulting tensor of cells used to represent the pointcloud. 
Point-based methods mainly use PointNet \cite{qi2017pointnet} or some deformation that introduces a deep network architecture to perform semantic segmentation or encode the raw point cloud data for 3D object detection, classification, and segmentation. These algorithms have less loss of spatial information of the environment because features are learned from the raw point cloud data, which makes their detection accuracy high. 
The PointNet methods can learn the rich spatial features of a raw point cloud, and the detection accuracies are usually high. However, their real-time performances are not ideal because of the high computational cost\cite{wang2020kda3d}.  
And Graph-based methods in which the points are converted into a graph, the points are considered nodes, and the connections of a point to its neighbors in a fixed radius are the edges. Graph representation is used to preserve the irregularity of the point cloud. Aggregating features along the edges iteratively update the vertex features. Point-GNN \cite{shi2020point}extracts features of the point cloud by iteratively updating vertex features on the same graph.
Other algorithms combine these two methods.PV-RCNN \cite{shi2020pv} combines voxelization and 3D sparse convolution with Keypoint pooling using the SA from PointNet. PP-RCNN \cite{tu2021pp} uses pillar feature extraction of the Pointpillars with keypoint sampling using the SA module from PointNet. Pyramid-RCNN \cite{mao2021pyramid} also uses this method.

\subsection{Feature Extraction Module}
This module employs various techniques to extract robust and high-level features from the structured representation of the LiDAR data. These techniques can be classified into two categories based on how the LiDAR data is represented. 2D Backbone Networks operate in 2D space and are mainly Fully Convolutional Networks (FCN). 2DBNs are applied to structured data in 2D space, such as pseudo-images, BEV, and FV range images, to extract rich, high-dimensional feature maps. The building blocks of a 2DBN are customarily made up of 2D convolution layers followed by a Batch Normalization\cite{ioffe2015batch} and ReLU\cite{nair2010rectified} operations. 
3D Backbone Networks operate in the 3D space and can be classified into three groups.
3D Convolutional Neural Networks (CNNs) are applied to structured data in 3D space, such as voxels. Due to the sparsity of LiDAR data, 3D CNNs are made up of 3D sparse convolution layers (3D SpConv) or 3D submanifold sparse convolution layers (3D Sub-SpConv) followed by a Batch Normalization and ReLU operation. SECOND\cite{yan2018second} implements the 3D SpConv and 3D Sub-SpConv to make a 3D backbone network for 3D object detection.
3DBN PointNet++ uses PointNet++ \cite{qi2017pointnet} segmentation form to extract both semantic segmentation scores and global context features for each point. PointNet++ consists of Set Abstraction layers to learn features and Feature Propagation layers to propagate the learned multi-scale features back to all points through interpolation.
3DBN Graph Neural Network is applied when the LiDAR data is represented as a graph. The GNN uses the aggregate of the features along the edges to refine the features at the vertex. The node features are extracted using Max operations and Multi-Layer Perceptrons(MLP), and the MLP weights are learned individually for each iteration. PointGNN\cite{shi2020point} utilizes this backbone network.

\subsection{Detection Module}
This module is utilized to make predictions on the pointcloud and can be categorized based on the approach method. Anchor-based networks can regress object location/size and output a classification score by searching through the feature map output of the backbone with anchors (boxes that define the shape and size of the objects of interest). Predictions can be made by picking objects whose size and location match the anchors. The most common anchor-based network architecture is the Region Proposal Network (RPN)\cite{ren2015faster}. RPN's input is the high-level feature map and outputs multiple 3D bounding box proposals, many of which will overlap and correspond to the same object. The redundant proposals are then removed using Non-Maximum Suppression (NMS)\cite{neubeck2006efficient}. Anchor-free networks make predictions by learning semantic binary label information for each point from the backbone and passing the information through fully connected layers to identify points or areas/parts that belong to an object. Then, for each point or part of an object, an object prediction is made, consisting of a class confidence score and a 3D bounding box.
Two-stage 3D object detectors include a prediction Refinement (PR) stage. A PR receives the proposed 3D bounding box from Detector Networks as input, samples it at a fine-grained level, and extracts features to improve the classification confidence score and the 3D bounding box location, size, and orientation.

The {\bfseries PointPillars} algorithm was chosen as the preferred LiDAR-based object detection method because of its favourable detection to computational complexity, as reported by the authors. It is able to achieve a fast inference of about 62 fps. This is possible because the 3D structure of the pointcloud is converted to a 2D representation and processed with a 2D CNN.

\begin{figure}[h!]
    \centering
    \includegraphics[scale = 0.30]{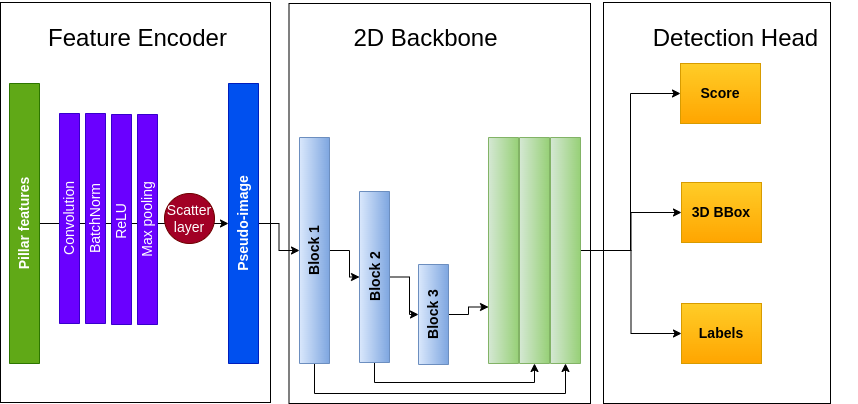}
    \captionsetup{justification=centering}
    \caption{\textrm{Overview of the PointPillars architecture}}
    \label{fig:2F}
\end{figure}
\section{PointPillars}
The PointPillars\cite{lang2019pointpillars} algorithm is a one-stage, grid-based 3D object detection algorithm whose input is a point cloud generated by a LiDAR sensor and estimates 3D bounding boxes for objects in the point cloud. The network is divided into three stages: The Pillar Feature Net converts a point cloud into a sparse pseudo-image, a 2D convolutional backbone that processes the pseudo-image and extracts high-level features, and a detection head that detects and regresses 3D boxes surrounding the objects.


The feature encoder module of the network discretizes the point cloud into a grid from the Bird's Eye View (BEV), and a set of pillars are generated. A 9-dimensional vector represents the points in these pillars  
   \[{\mathbf {D}}= [x,y,z,r,X_c,Y_c,Z_c,X_p,Y_p].\] 
The subscripts c and p denote offsets from the centroid of the pillars and offsets from the geometric center of the pillar in the x-y plane, respectively. A simplified version of PointNet is used to create a tensor ({\bfseries C, H, W}), which is the pseudo-image where C is the content of each cell of the pseudo-image, and H and W indicate the height and width of the image.
   
A 2D CNN backbone is used to extract features from the pseudo-image. It consists of sequential convolution layers, which are downsampled and subsequently upsampled to learn features at different scales. Batch Normalization and ReLU operations are applied at every convolution layer. The final feature map is obtained by concatenating all up-sampled output pillar feature maps.

The Detection head uses a single shot detector \cite{liu2016ssd}. The SSD matches the predicted anchor boxes to the ground truth using 2D Intersection over Union(IOU), and a Non-Maximum suppression is then applied to filter noisy predictions

\subsection{Data Collection and Processing}
Tunnel navigation and UAV flight tests were conducted using the setup described in the previous section, and pointcloud scans of the scene were recorded using the LiDAR sensor mounted on the UGV. The recorded pointclouds constitute the dataset used for the training and evaluation of the network. The pointclouds are extracted from the recorded ROS bags using the PCL library and are saved in PCD format. The pointclouds are unorganized, with an average of 120000 points per scan.
The dataset annotation was achieved using the MATLAB LidarLabeler tool. The object to be labeled is bounded by a cuboid that shrinks just enough to fit all the points in the cluster that forms the object. The cuboid model parameters are stored as a nine-element row vector of the form \([x_{ctr},y_{ctr},z_{ctr},x_{len},y_{len},z_{len},x_{rot},y_{rot},z_{rot}]\) \cite{noauthor_parametric_nodate}. The subscripts \(ctr,len, rot\) represent the center, length, and rotation in the corresponding plane, respectively. 6660  annotated pointclouds were obtained from the  20174 pointclouds originally extracted. This was because the UAV was not visible in most of the pointclouds.

Due to the high detection range of the Velodyne sensor, \(\approx\) 220m. Sections of the pointcloud include points outside the boundaries of the tunnel. The pointcloud is cropped to cover only the area covered by the tunnel using adjusted parameters defined by \cite{lang2019pointpillars}. The pointcloud is discretized into a grid using the following parameters. 


\[X_{min} = 0_m , Y_{min} = -39.68_m , Z_{min}= -7.0_m\]
\[X_{max} = 70_m , Y_{max} = 39.68_m , Z_{max}= 5.0_m\]
\[X_{step} = 0.16_m , Y_{step} = 0.16_m\]

The min and max subscripts denote the minimum and the maximum distances along the x,y, and z axes, respectively. Step is the resolution along the x and y axes, respectively.


The dimensions of the pseudo-image are calculated using the following equations.
\begin{equation}
	    X_{n}= \Bigg\lfloor \left(\dfrac{X_{max}-X_{min}}{X_{step}}\right) \Bigg\rceil
\end{equation}
\begin{equation}
    	Y_{n}= \Bigg\lfloor \left(\dfrac{Y_{max}-Y_{min}}{Y_{step}}\right) \Bigg\rceil
\end{equation}

Finally, the pillar extraction parameters are defined as

       
\[
\text{gridParams} =
\begin{bmatrix}
X_{\min},\ Y_{\min},\ Z_{\min} \\
X_{\max},\ Y_{\max},\ Z_{\max} \\
X_{\text{step}},\ Y_{\text{step}},\ ds_{\text{factor}} \\
X_n,\ Y_n,\ \_
\end{bmatrix}
\]

The data set is then split into training and test sets. We selected a 75\%-25\% split for the training and testing data, respectively, because our data set is small and selecting more data for training increases the model's accuracy.
The table \ref{tb:config} details the software and hardware used in the training process.

\begin{table}[!htb]
\centering
\caption{ Configuration used for training and testing}
\begin{tabular}{llllll} 
\hline
Software and hardware configuration    & Version of the model            \\ 
\hline
Processor                            & Intel Core i7-10700K @ 2.90Ghz  \\
Graphics Card                        & GeForce GTX 1660 SUPER          \\
RAM                                  & 32GB                             \\
Programming Language                 & MATLAB                    \\
\hline
\end{tabular}\label{tb:config}
\end{table}

A dlnetwork object is created that takes in a mini-batch of data pillarFeatures and pillarIndices with corresponding ground truth boxes, anchor boxes, and grid parameters as the input. The network was trained using the following hyperparameters: mini-batch size 2, learning rate \(2*10^{-4}\), learnDropRatePeriod 15, learnRateDropFactor 0.8, gradient decay factor 0.9 at 70, and 150 epochs. The network uses the ADAM optimizer for end-to-end training.

\section{Results and Discussion}
\begin{figure*}[h]
\centering
\includegraphics[scale = 0.7, width=0.43\linewidth]{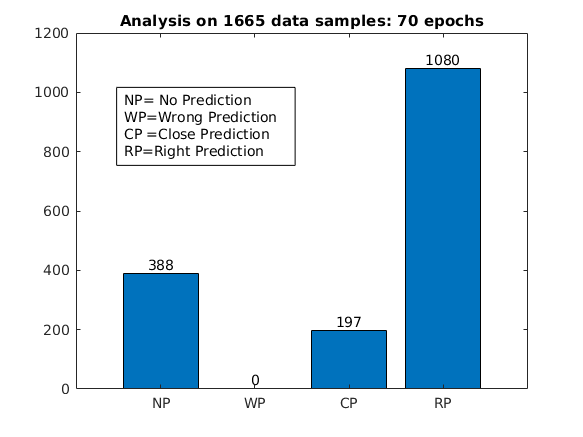}
\includegraphics[scale = 0.7, width=0.43\linewidth]{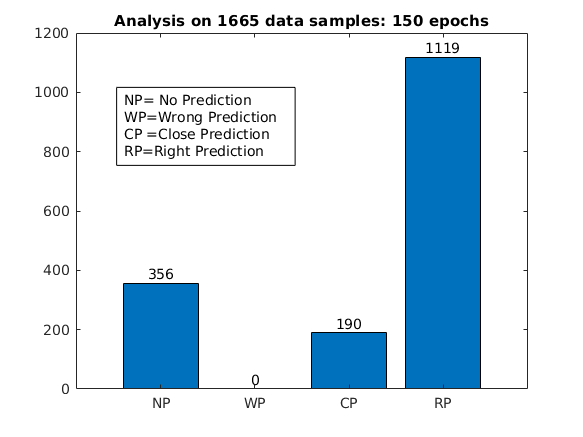}
\caption{Detection results of the network trained for 70 epochs (left); 150 epochs (right)}
\label{fig:4B}
\end{figure*}

The first test the network was subjected to evaluated its prediction accuracy by comparing its solution to the ground truth annotation on the test dataset. Figure \ref{fig:4B} shows the detection results of the network trained for 70 epochs.
To evaluate how accurate a prediction is, we use a heuristic {\bfseries d}, which is the 3D position error.
\begin{equation}
	   \mathbf{d} = \sqrt{(x_b - x_t)^2 + (y_b - y_t)^2 + (z_b - z_t)^2}
\end{equation}
The subscripts b and t represent the predicted bounding box center and the annotated bounding box center. 

The following notations are used to describe the detection result.

\begin{itemize}
    \item NP: No Prediction, this is a case where the network is unable to make a prediction on the pointcloud, thus returns no bounding box.
    \item WP: Wrong Prediction, the network predicts a bounding box but the bounding box is not around the points representing the UAV i.e {\bfseries d} is greater than 40cm.
    \item CP: Close prediction, the network predicts a bounding box and {\bfseries d} is greater than 20 cm but less than 40 cm. This is to accommodate cases where the prediction of the elevation is off.
    \item RP: Right prediction, the predicted bounding box aligns with the annotated box and {\bfseries d} is less than 20cm
\end{itemize}

The network was also trained for 150 epochs to observe if there are benefits to training it for longer. As shown in figure \ref{fig:4B}, training for longer epochs only produces an increase of 2.4\%, moving the number of correct predictions from 1277 to 1309.

To evaluate the network's performance, a dataset of unlabeled pointclouds from the recorded data was used. This data contains the VICON solution used to obtain the approximate location of the UAV to the UGV, the LiDAR position estimation solution obtained from the clustering method, and also the overall EKF solution obtained by fusing all the sensor data. The VICON solution returns the relative position of the reflective marker placed on the UAV's IMU with respect to the marker affixed to the top of the LiDAR sensor on the UGV. Several transforms were applied to align the EKF solution to the reference VICON solution, as explained in \cite{gross2019field}. The solution produced by the network does not account for these transforms, as shown in figure \ref{fig:iters} below.

It can be observed in figure \ref{fig:iters} at first glance that the network solutions provided more updates than the clustering method. The network produces 265 updates \(\approx\) 4X more than the 60 updates produced by the clustering method. Both methods start making position estimates around the 60-second mark because that is when the UAV enters the field of view of the LiDAR sensor. As long as the UAV is in the Field of View (FOV) of the LiDAR sensor, the network can detect the drone and return the relative position with centimeter-level errors in the X and Y-axis. The solution deviates for the Z-axis, showing meter-level errors. 

\begin{table}[htb!]
\caption{ Position error statistics of the LiDAR estimates}
\centering
\begin{tabular}{llllll}
\hline
                            &      & RMS(m) & mean(m) & std(m) & Max(m) \\ \hline
{Clustering} & $X_{err}$ & 0.22   & 0.18    & 0.13   & 0.47   \\
                            & $Y_{err}$ & 0.17   & 0.15    & 0.08   & 0.35   \\
                           & $Z_{err}$ & 0.23       & 0.14        & 0.18       & 1.15\\ \hline
{Network}    & $X_{err}$ & 0.27   & 0.26    & 0.06   & 0.39   \\
                            & $Y_{err}$& 0.20   & 0.19    & 0.05   & 0.29   \\ 
                           & $Z_{err}$  & 2.42       & 2.42        & 0.15       & 2.63       \\ \hline
\end{tabular}\label{tb:Perr}
\end{table}

As shown in Table \ref{tb:Perr},  the X and Y position estimate errors produce similar values to that of the clustering method, with only a difference of 5 cm in the Root Mean Square (RMS) of the errors. However, the RMS of the position estimate error in the Z-axis is blown up to 2.42 m. 

\begin{figure}[!h]
    \centering
    \includegraphics[scale = 0.8, width=0.99\linewidth]{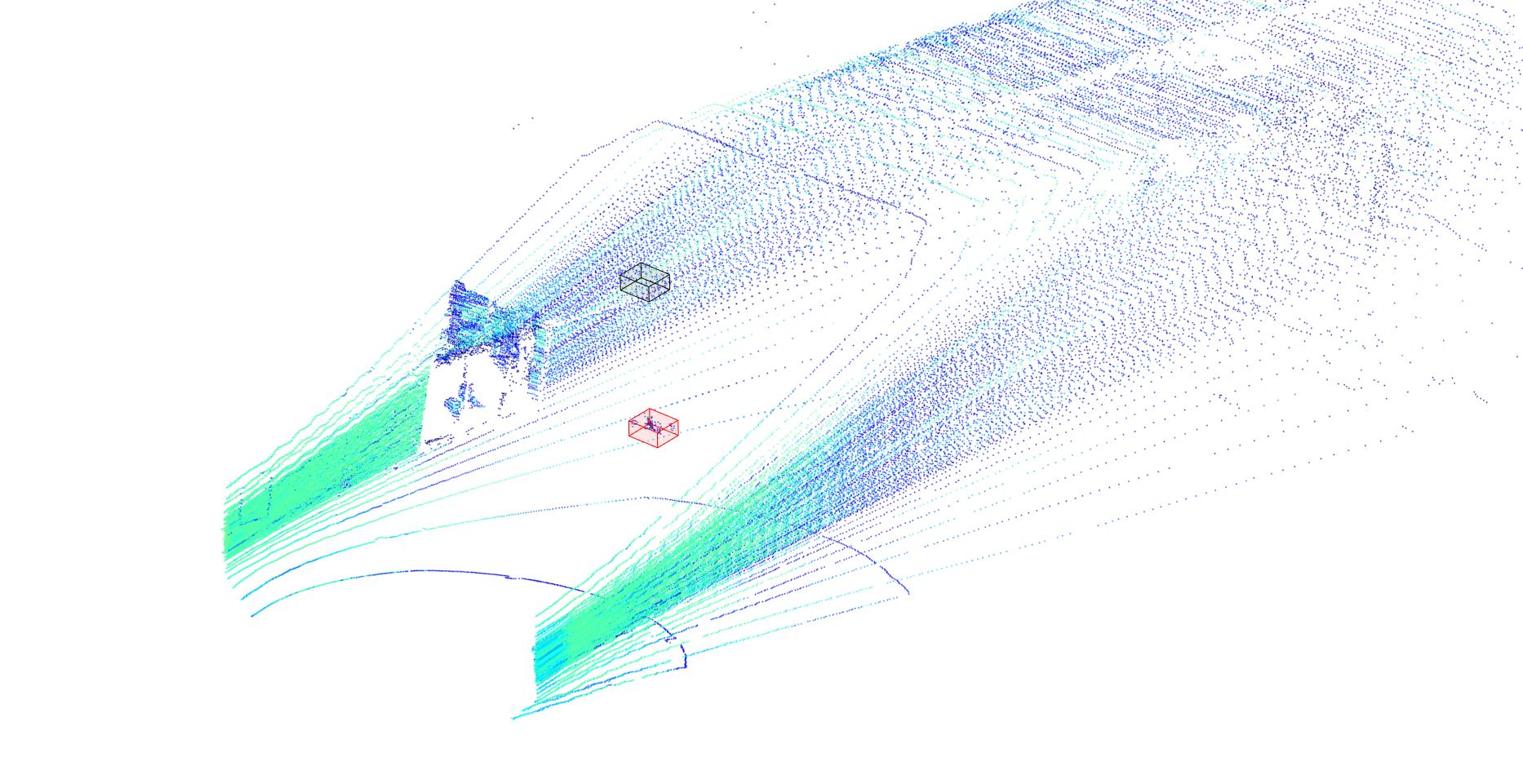}
    \caption{\textrm{Detection result on pointcloud at a single epoch (black bounding box, VICON estimate; red bounding box, Network estimate)}}
    \label{fig:4D}
\end{figure}
\begin{figure}[!h]
    \centering
    \includegraphics[scale = 0.8, width=0.99\linewidth]{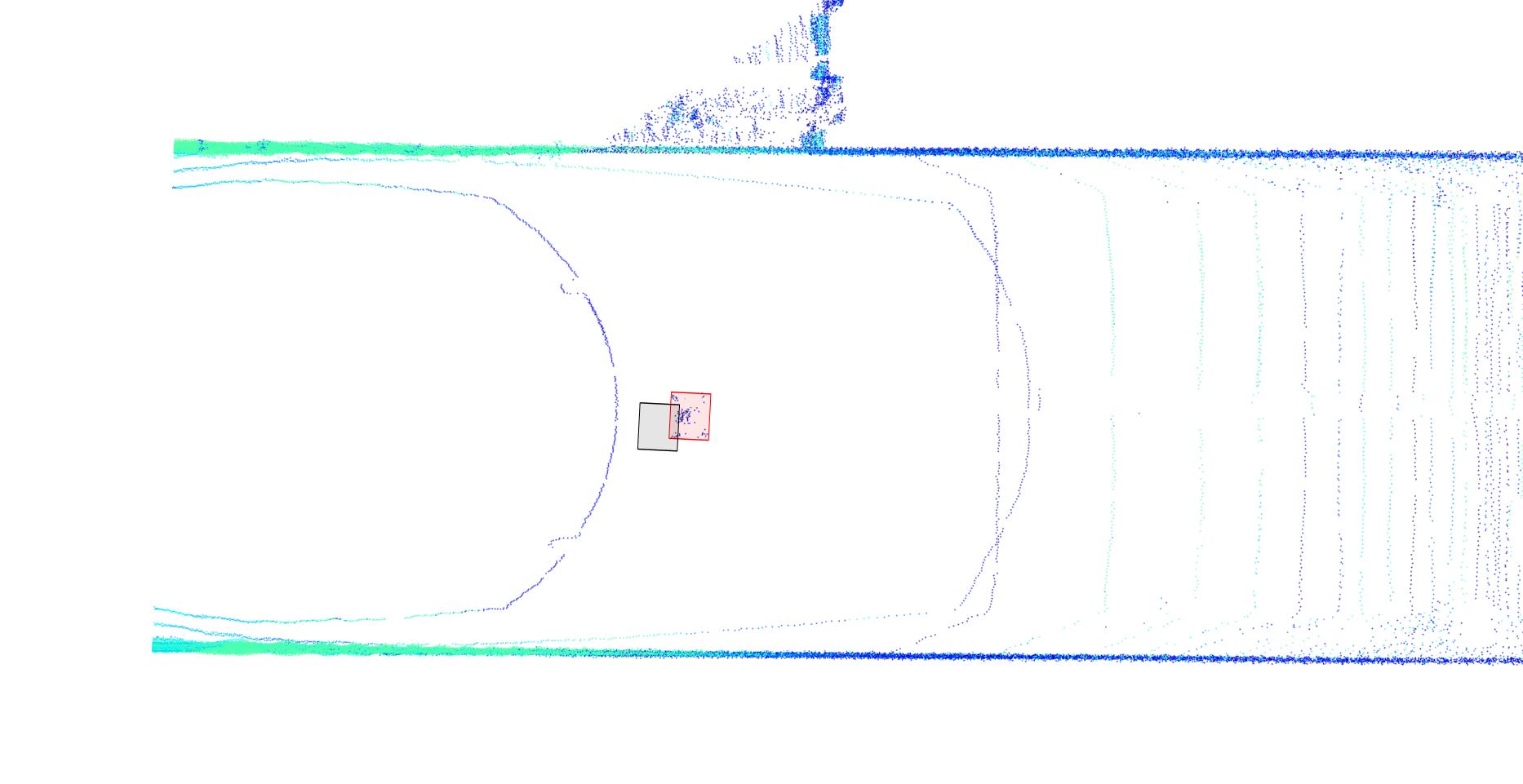}
    \caption{\textrm{Top view of the pointcloud at a single epoch (black bounding box, VICON estimate; red bounding box, Network estimate)}}
    \label{fig:4J}
\end{figure}

One postulate as to why this occurs is that the transforms employed by the system in real-time are not applied to the pointclouds in the recorded data. 
Figure \ref{fig:4D} and \ref{fig:4J} show the position of the VICON solution in the black bounding box and the network prediction in the red bounding box, which by visual inspection cannot be the true position of the UAV in the pointcloud. This error propagates the entirety of the solution, as shown in figure \ref{fig:iters}. The Z position of the UAV was observed to be displaced by a factor of 2.4, as evident by the error analysis. 

\begin{figure*}[h!]
     \centering
     \includegraphics[scale = 0.30]{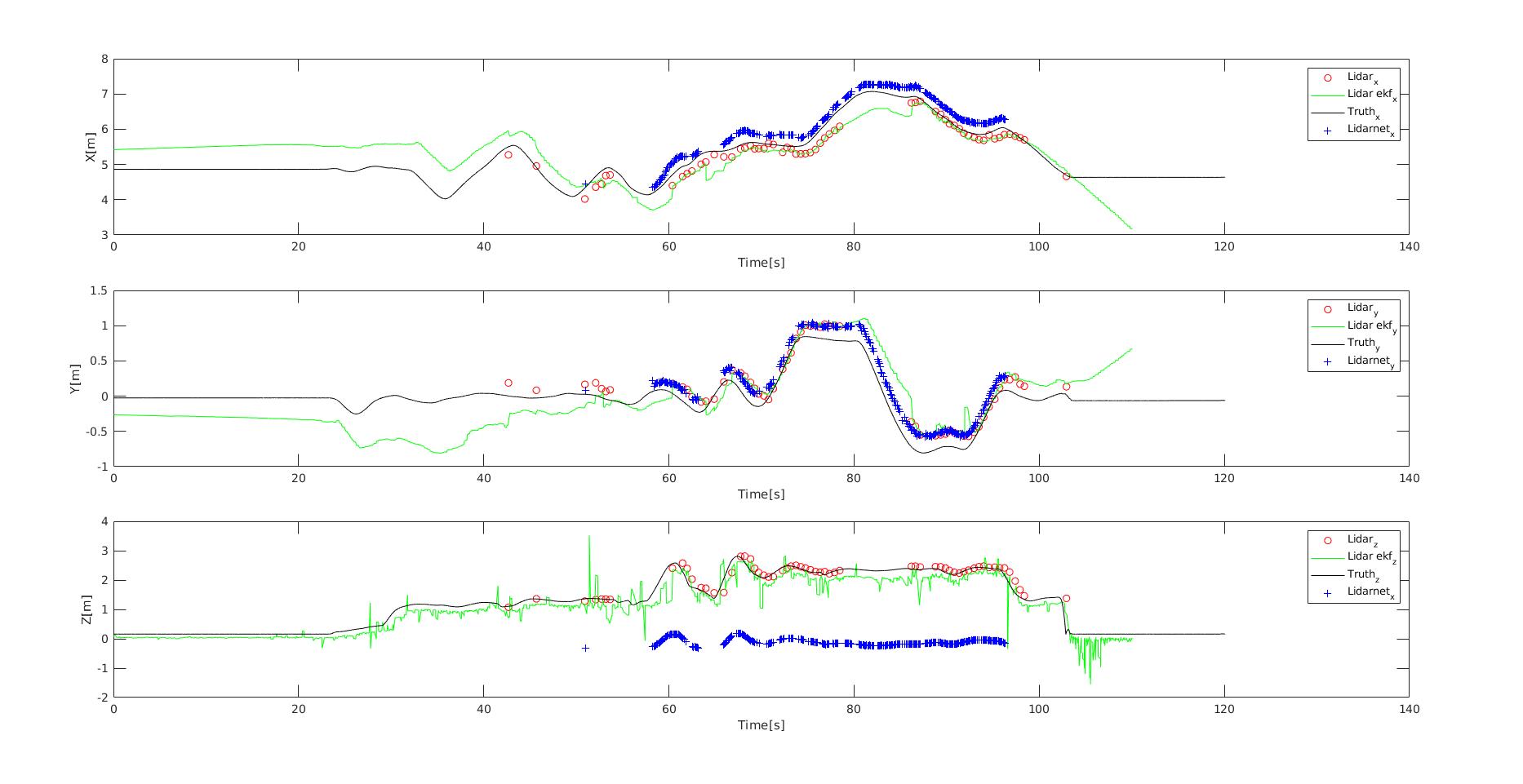}
     \caption{Pointcloud representation of the tunnel with the UAV bounded by the green box}
     \label{fig:iters}
\end{figure*}

To correct for the error observed in the Z-axis, the transforms used in the system in real-time would need to be applied to the pointclouds. But these transforms were not recorded in the data file, thus only abstract corrections can be made at the time of this research. If the observed position estimation error were applied to the network's solution, the overall solution aligns with the truth reference solution as shown in the error statistics in Table \ref{tb:Zerr}

\begin{table}[!htb]
\caption{ Corrected Z-error statistics of the LiDAR estimates}
\centering
\begin{tabular}{llllll}
\hline
              &      & RMS(m) & Mean(m) & Std(m) & Max(m) \\ \hline
Clustering     & $Z_{err}$ & 0.23   & 0.14    & 0.18   & 1.15   \\
Network + 2.42 & $Z_{err}$ & 0.15   & 0.11    & 0.11   & 0.75   \\ \hline
\end{tabular}\label{tb:Zerr}
\end{table}
Table \ref{tb:3Derr} shows the positioning error statistics for the LiDAR position updates from the clustering method compared with the object detection method. The LiDAR updates are not synchronous with the reference truth; thus, the solution at the closest epoch is chosen for the error analysis. The 'T' here is the corrected Z position estimate.

\begin{table}[htbp]
\addtolength{\tabcolsep}{-4pt}
\caption{ 3D Positioning error statistics of the LiDAR estimates}
\begin{tabular}{llllllll}
\hline
         & \# of updates & \# of Scans & RMS(m) &  Mean(m) &  Std(m)  & Max(m) \\  \hline
Clustering     & 60            & 1231         & 0.35   & 0.34    & 0.09      & 1.16   \\
Network   & 265           & 1231         & 2.44   & 2.44    & 0.16       & 2.66   \\
Network+'T' & 265           & 1231         & 0.37   & 0.36    & 0.06       & 0.76   \\ \hline
\end{tabular}\label{tb:3Derr}
\end{table}

The RMS of the network's solution is dominated by a mean of \(\approx\) 2.44 m and a standard deviation of \(\approx\)16 cm, which is relatively small. The RMS shows how much the solution deviates from the truth and, in this case, relates back to the observed transform in the solution. If the solution is corrected with the visually observed 'transform,' error analysis of the network's solution shows results similar to those of the clustering method. 
One thing to note is that the network returns the relative position of the center of the bounding box to the center of the LiDAR sensor. In contrast, the reference truth solution returns the position of the marker on the UAV to the marker on the LiDAR sensor.

\section{Conclusion and Future work}
This paper has presented an implementation of the PointPillars algorithm in MATLAB for UAV position estimation utilizing object detection in a 3D space. The implemented network was trained and tested with data captured during experiments detailing the flight sequence in a collaborative UAV/UGV operation conducted by the WVU Navlab. The results show that we can detect the UAV in the point cloud and return its relative position to the UGV. In the test performed to evaluate the accuracy of the network, it can be observed that the network made no wrong predictions. The absence of wrong predictions could be because the processed pointcloud had no other objects with a semblance to the shape of the UAV; thus, no false negatives were produced. With our dataset, there's no way to guarantee that the accuracy of the network is preserved. A suggested solution would be to conduct tests in a cluttered environment and then train the network with the data obtained. 

An observed setback is that the transforms employed in the system configuration are not transferred to the recorded data. This does not seem like a problem for online/real-time testing, but it becomes very obvious in the case of offline tests. As shown in figure \ref{fig:4D}, the network was able to detect the UAV accurately, but the corresponding truth data returned a relative position different from the position of the UAV in the pointcloud. This complication can be mitigated by implementing the applied transforms to the recorded data. Table \ref{tb:3Derr} highlights the network's better detection performance than the clustering method. The network could produce 4X more correct predictions than the clustering method. The network solution rivals the solution of the clustering method, which means the position estimation can be achieved with the network in real time. Therefore, we can remove the altimeter and the UWB ranging radio from the drone payload as they become redundant.

Future work would investigate how a refinement stage can be incorporated into the network architecture for better detection accuracy. The network would be implemented using the NVIDIA TensorRT to achieve the reported inference time in real-time testing. It would also be an excellent approach to implement and observe how other 3D detection algorithms compare to the implemented PointPillars for this use case. Finally, as with any deep learning model, more training and parameter tuning are required to improve the model's results.

\bibliographystyle{IEEEtran}
\bibliography{references}
\vfill
\end{document}